\documentclass[11pt,a4paper]{article}
\usepackage[hyperref]{naaclhlt2018}
\usepackage{times}
\usepackage{latexsym}
\usepackage{graphicx}
\usepackage{multirow}

\usepackage{url}

\aclfinalcopy 

\title{Knowledge-enriched Two-layered Attention Network for Sentiment Analysis}

\author{Abhishek Kumar$^a$, Daisuke Kawahara$^b$, Sadao Kurohashi$^b$ \\
  $^a$Indian Institute of Technology Patna, India\\
  $^b$Kyoto University, Japan\\
  {\tt \{abhishek.ee14\}@iitp.ac.in} \\
  {\tt \{dk,kuro\}@i.kyoto-u.ac.jp}}

\date{}

\begin{document}
\maketitle
\begin{abstract}
We propose a novel two-layered attention network based on Bidirectional Long Short-Term Memory for sentiment analysis. The novel two-layered attention network takes advantage of the external knowledge bases to improve the sentiment prediction. It uses the Knowledge Graph Embedding generated using the WordNet. We build our model by combining the two-layered attention network with the supervised model based on Support Vector Regression using a Multilayer Perceptron network for sentiment analysis. We evaluate our model on the benchmark dataset of SemEval 2017 Task 5. Experimental results show that the proposed model surpasses the top system of SemEval 2017 Task 5. The model performs significantly better by improving the state-of-the-art system at SemEval 2017 Task 5 by 1.7 and 3.7 points for sub-tracks 1 and 2 respectively.
\end{abstract}

\section{Introduction}

With the rise of microblogging websites, people have access and option to reach to the large crowd using as few words as possible. Microblog and news headlines are one of the common ways to dispense information online. The dynamic nature of these texts can be effectively used in the financial domain to track and predict the stock prices \cite{goonatilake2007volatility}. These can be used by an individual or an organization to make an informed prediction related to any company or stock \cite{si2013exploiting}.

This gives rise to an interesting problem of sentiment analysis in financial domain. A study indicates that sentiment analysis of public mood derived from Twitter feeds can be used to eventually forecast movements of individual stock prices \cite{smailovic2014stream}. An efficient system for sentiment analysis is a core component of a company involved in financial stock market price prediction. 

Social media texts are prone to word shortening, exaggeration, lack of grammar and appropriate punctuations. Moreover, the word limit constraint forces a user to limit their content and squeeze in their opinion about companies. These inconsistencies make it challenging to solve any natural language processing tasks including sentiment analysis \cite{khanarian}.

Bag-of-words and named entities were used by \newcite{999abb56de63450f86fb01e488b5566c} for predicting stock market. For predicting the explicit and implicit sentiment in the financial text, \newcite{VandeKauter20154999} used a fine-grained sentiment annotation scheme. \newcite{kumar2017iitpb} used a classical supervised approach based on Support Vector Regression for sentiment analysis in financial domain. \newcite{OliveiraCA13} relied on multiple regression models.  \newcite{akhtar2017multilayer} used an ensemble of four different systems for predicting the sentiment. It used a combination of Long Short-Term Memory (LSTM) \cite{lstm1997}, Gated Recurrent Unit (GRU) \cite{Cho-GRU}, Convolutional Neural Network (CNN) \cite{cnn2014} and Support Vector Regression (SVR) \cite{svr:2004}. \newcite{yang2016hierarchical} used a hierarchical attention network to build the document representation incrementally for document classification.

Our model focuses on interpretability and usage of knowledge bases. Knowledge bases have been recognized important for natural language understanding tasks \cite{Minsky:1986:SM:22939}. Our main contribution is a two-layered attention network which utilizes background knowledge bases to build good word level representation at the primary level. The secondary attention mechanism works on top of the primary layer to build meaningful sentence representations. This provides a good intuitive working insight of the attention network.

\section{Proposed Methodology}

\begin{figure}[h!]
\centering
\includegraphics[width=0.32\textwidth]{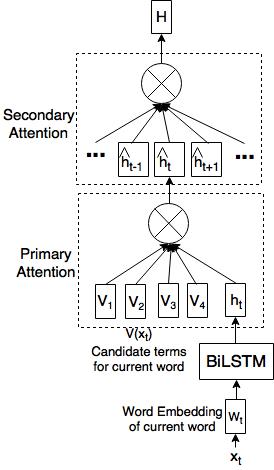}
\caption{Two-layered attention network}
\label{fig:model}
\end{figure}

\begin{figure}[h!]
\centering
\includegraphics[width=0.33\textwidth]{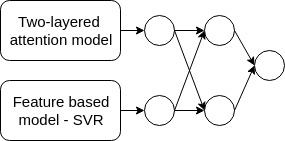}
\caption{Multilayer perceptron based ensemble}
\label{fig:ensemble}
\end{figure}

We propose a two-layered attention network which leverages external knowledge for sentiment analysis. It consists of a bidirectional Long Short-Term Memory (BiLSTM) \cite{Graves:2005:BLN:1986079.1986220} based word encoder, word level attention mechanism for capturing the background knowledge and a sentence level attention mechanism aimed at grasping the context and the important words. The output of the two-layered attention network is then ensembled with the output of the feature based SVR using the Multilayer perceptron based approach described in \newcite{akhtar2017multilayer}. The overall ensembled system is shown in Figure \ref{fig:ensemble}. Each of the components is explained in the following subsections and an overview of the two-layered attention network is depicted in Figure \ref{fig:model}.

\subsection{Two Layered Attention Model}

\subsubsection{BiLSTM based word encoder}

A Long-Short Term Memory (LSTM) is a special kind of Recurrent Neural Network. It handles the long-term dependencies where the current output is dependent on many prior inputs. BiLSTM, in essence, is a combination of two different LSTM - one working in forward and the other working in the backward direction. The contextual information about both past and future helps in determining the current output. 

The two hidden states \overrightarrow{h_t} and \overleftarrow{h_t} for forward and backward LSTM are the information about past and future respectively at any time step $t$. Their concatenation $h_t$ = [\overrightarrow{h_t}, \overleftarrow{h_t}] provides complete information. Each word of the sentence is fed to the network in form of word embeddings which are encoded using the BiLSTM.

\subsubsection{Word Level Attention}

External knowledge in form of Knowledge Graph Embedding \cite{yang15:embedding} or top-k similar words are captured by using the word level attention mechanism. This serves the purpose of primary attention which leverages the external knowledge to get the best representation for each word. At each time step we get V($x_t$) relevant terms of each input $x_t$ with $v_i$ being the embedding for each term. (Relevant terms and embeddings are described in next section). The primary attention mechanism assigns an attention coefficient to each of relevant term having index $i$ $\in$ V($x_t$):

\begin{equation}
\alpha_{ti} \propto h_{t}^TW_{v}v_i\label{eq:1}
\end{equation}

\noindent where $W_v$ is a parameter matrix to be learned.

\begin{equation}
m_t = \sum_{i\in{V(x_t)}}\alpha_{ti}v_i\label{eq:2}
\end{equation}

\begin{equation}
\widehat{h_t} = m_t+h_{t}\label{eq:3}
\end{equation}

The knowledge aware vector ($m_t$) is calculated as Equation \ref{eq:2}, which is concatenated with the hidden state vector to get the final vector representation for each word.

\subsubsection{Sentence Level Attention}

The secondary attention mechanism captures important words in a sentence with the help of context vectors. Each final vector representing the words is assigned a weight indicating its relative importance with respect to other words. The attention coefficient $\alpha_t$ for each final vector representation is calculated as:

\begin{equation}
\alpha_{t} \propto \widehat{h_{t}^T}W_{s}u_s\label{eq:4}
\end{equation}

\begin{equation}
H = \sum_{t}\alpha_{t}\widehat{h_{t}}\label{eq:5}
\end{equation}

\noindent where $W_s$ is a parameter matrix and $u_s$ is the context vector to be learned. $H$ is finally fed to a one layer feed forward neural network.

\subsection{Relevant Terms and Embeddings}

External knowledge can provide explicit information for the model which the training data lacks. This helps the model to make better predictions. We relied on Knowledge Graph Embeddings based on WordNet and Distributional Thesaurus to get relevant terms and their corresponding embeddings for each word in the text.

\subsubsection{Knowledge Graph Embedding}

WordNet\footnote{https://wordnet.princeton.edu} is a lexical database which contains triplets in the form of (subject, relation, object). Both subject and object are synsets in WordNet. Each word in the text serves as the subject of the triplet. The relevant terms for the current word are the triplets having the current word as the subject. We then employ Knowledge Graph Embeddings to learn the representation of the triplet. A 100-dimensional dense vector representation for each subject, relation and object were learned using the DistMult approach \cite{yang15:embedding} and concatenated. These served as the relevant embeddings. An example of triplet in WordNet is (\textit{bronze\_age}, \textit{part\_of}, \textit{prehistory}).

\subsubsection{Distributional Thesaurus}

Distributional Thesaurus (DT) \cite{DBLP:journals/jlm/BiemannR13} is an automatically computed word list which ranks words according to their semantic similarity. We use a pre-trained DT to expand a current word. For each current word, top-4 target words are found which are the relevant terms. The relevant embeddings are obtained by using a 300-dimensional pre-trained Word2Vec \cite{mikolov2013distributed} and GloVe \cite{pennington2014glove} model. An example of the DT expansion of the word 'touchpad' is \textit{mouse}, \textit{trackball}, \textit{joystick} and \textit{trackpad}.

\subsection{Feature Based Model - SVR}

The following hand-crafted features are extracted and used to train a Support Vector Regression (SVR).

\noindent \textbf{- Tf-Idf:} Term frequency-inverse document frequency (Tf-Idf) reflects the importance of each word in a document. We use Tf-Idf score as a feature value for each word.

\noindent \textbf{- Lexicon Features:} Sentiment lexicons are an important resource for sentiment analysis. We employ the following lexicons: Bing Liu opinion lexicon \cite{ding2008holistic} and MPQA subjectivity lexicon \cite{mpqa}, SentiWordNet \cite{baccianella2010sentiwordnet} and Vader sentiment \cite{vader2014}. From the above lexicons we extracted the agreement score \cite{rao2012analyzing} and the count of the number of occurrences of all positive and negative words in the text.

\noindent \textbf{- Word embedding:} We use the 300-dimensional pre-trained Word2Vec and GloVe embedding. The sentence embedding was obtained by concatenating the embedding for all words in the sentence.

\section{Experiments}

\subsection{Dataset}

We evaluate our proposed approach for sentiment analysis on the benchmark datasets of SemEval-2017 shared task 5. The task 'Fine-Grained Sentiment Analysis on Financial Microblogs and News' \cite{semeval20175} had two sub-tracks. Track 1 - 'Microblog Messages' had 1,700 and 800 train and test instances respectively. Track 2 - 'News Statements \& Headlines' had 1,142 and 491 train and test instances respectively. The task was to predict a regression score in between -1 and 1 indicating the sentiment with -1 being negative and +1 being positive.

\subsection{Implementation Details}

We implement our model using Tensorflow and Scikit-learn on a single GPU. We use a single layer BiLSTM with the two-layered attention mechanism followed by a one layer feed forward neural network. The number of units in each LSTM cell of the BiLSTM was 150. The batch size was 64 and the dropout was 0.3 \cite{dropout} with the Adam \cite{adam} optimizer. The length of context vector in the secondary attention network was 300. For each experiment, we report the average of five random runs. Cosine similarity is a measure of similarity. It represents the degree of agreement between the predicted and gold values. Cosine similarity was used for evaluation as per the guideline.

\subsection{Results}

We compare our system with the state-of-the-art systems of SemEval 2017 Task 5 and the system proposed by \newcite{akhtar2017multilayer}. Table \ref{result-test} shows evaluation of our various models. Team ECNU \cite{ecnu2017} and Fortia-FBK \cite{mansar2017fortia} were the top systems for sub-tracks 1 and 2 respectively. Team ECNU and Fortia-FBK reported a cosine similarity of 0.777 and 0.745 for sub-tracks 1 and 2 respectively. Team ECNU employed a number of systems - Support Vector Regression, XGBoost Regressor, AdaBoost Regressor and Bagging Regressor ensembled together. Team Fortia-FBK used a Convolutional Neural Network for this task. The system proposed by \citeauthor{akhtar2017multilayer} utilizes an ensemble of LSTM, GRU, CNN and a SVR and reported a cosine similarity of 0.797 and 0.786 for the two sub-tracks.

Our proposed system has a cosine similarity of 0.794 and 0.782 for sub-tracks 1 and 2 respectively. The proposed system performs significantly better than top systems of SemEval 2017 Task 5 for both the tasks. Moreover, the system performs at par with the system proposed by \citeauthor{akhtar2017multilayer} with half the number of subsystems involved in the ensemble. This shows that our proposed system is not only robust since it performs for both the task equally well but also powerful as it involves fewer subcomponents while having the same expressive power. 

The two-layered attention network alone performs better than the best system of SemEval 2017 Task for both the sub-track. It manages to achieve much higher score than any of the deep learning component utilized by the system proposed by \newcite{akhtar2017multilayer} as shown in Table \ref{result-test-comparison-new}. This shows that the two-layered attention network helps to reduce overall model complexity without compromising the performance.

\begin{table}[h!]
\begin{center}
\resizebox{0.48\textwidth}{!}{
\begin{tabular}{|l|l|c|c|}
\hline 
& \bf Models & \bf Microblog & \bf News\\ \hline \hline
\multicolumn{4}{|l|}{Layered Attention Network} \\ \hline
L1 & Knowledge Graph Embedding & 0.758 & 0.727 \\
L2 & Distributional Thesaurus + GloVe  & 0.764 & 0.749 \\
L3 & Distributional Thesaurus + Word2Vec &  0.779 &  0.763 \\ \hline
\multicolumn{4}{|l|}{Support Vector Regression} \\ \hline
S1 & Tf-Idf + Lexicon & 0.735 & 0.720 \\ 
S2 & Tf-Idf + Lexicon + GloVe &  0.755 &  0.753 \\
S3 &Tf-Idf + Lexicon + Word2Vec & 0.743 & 0.740 \\ \hline \hline
\multicolumn{4}{|l|}{ Ensemble} \\ \hline
E1  & L3 + S2 &  0.794 &  0.782 \\ \hline
\end{tabular}}
\end{center}
\caption{Cosine similarity score of various models on test dataset.}
\label{result-test}
\end{table}

\begin{table}[h!]
\begin{center}
\resizebox{0.48\textwidth}{!}{
\begin{tabular}{|l|c|c|}
\hline 
Models & \bf Microblog & \bf News\\ \hline \hline
\multicolumn{3}{|l|}{Single systems} \\ \hline
\citeauthor{mansar2017fortia} (Team Fortia-FBK)  & - & 0.745 \\
\citeauthor{akhtar2017multilayer} - LSTM &  0.727 &  0.720 \\
\citeauthor{akhtar2017multilayer} - GRU &  0.721 &  0.721 \\
\citeauthor{akhtar2017multilayer} - CNN &  0.724 &  0.722 \\
L3 (proposed) &  0.779 &  0.763 \\ \hline
\multicolumn{3}{|l|}{Ensembled systems} \\ \hline
\citeauthor{ecnu2017} (Team ECNU) & 0.777 & 0.710 \\
\citeauthor{akhtar2017multilayer} & 0.797 & 0.786 \\
E1 (proposed) & 0.794 & 0.782 \\ \hline
\end{tabular}}
\end{center}
\caption{Comparison with the state-of-the-art systems.}
\label{result-test-comparison-new}
\end{table}

\subsection{Error Analysis}

We performed error analysis and observed that the proposed system faces difficulty at times. Following are the situations when the system failed and incorrectly predicted values of the opposite polarity:

\noindent $\bullet$ Sometimes the system fails to identify an intensifier. In the example below, 'pure' is used as an intensifier. \\
\textbf{Text :} Pure garbage stock \\
\textbf{Actual:} -0.946 \indent \textbf{Predicted:} 0.042

\noindent $\bullet$ The system fails when it does not have enough real-world information. In the example below, a low share price is a good opportunity to buy for an individual but from a company's point of view, a low share price does not indicate a prosperous situation.  \\
\textbf{Text :} Good opportunity to buy \\
\textbf{Actual:} -0.771 \indent \textbf{Predicted:} 0.260

\section{Conclusion}

In this paper, we proposed an ensemble of a novel two-layered attention network and a classical supervised Support Vector Regression for sentiment analysis. The two-layered attention network has an intuitive working. It builds the representation hierarchically from word to sentence level utilizing the knowledge bases. The proposed system performed remarkably well on the benchmark datasets of SemEval 2017 Task 5. It outperformed the existing top systems for both the sub-tracks comfortably. Experimental results demonstrate that the system improves the state-of-the-art system of SemEval 2017 Task 5 by 1.7 and 3.7 points for sub-tracks 1 and 2 respectively. This robust system can be effectively used as a submodule in an end-to-end stock market price prediction system. 

\section{Acknowledgements}

This work was supported by the Kyoto University Cooperative Intelligence Short-term Internship Program.

\bibliography{naaclhlt2018}
\bibliographystyle{acl_natbib}

\appendix

\end{document}